\documentclass{new_tlp}
\pdfoutput=1

\title[Theory and Practice of Logic Programming]
{A Distributed Approach to LARS Stream Reasoning (System paper)}

\author[T. Eiter, P. Ogris, K. Schekotihin]
{
	THOMAS EITER\textsuperscript{1}, PAUL OGRIS\textsuperscript{2}, and
	KONSTANTIN SCHEKOTIHIN\textsuperscript{2}\\
	\textsuperscript{1} Technische Universität Wien, Institut für Logic and Computation, KBS Group,\\
	\email{eiter@kr.tuwien.ac.at}\\
	\textsuperscript{2} Alpen-Adria-Universität, Klagenfurt, Austria,\\
	\email{\{konstantin.schekotihin,paul.ogris\}@aau.at}
}

\jdate{September 2019}
\pubyear{2019}
\pagerange{\pageref{firstpage}--\pageref{lastpage}}

\usepackage{amssymb}
\usepackage{amsmath}
\usepackage{graphicx}
\graphicspath{{figures/}}

\usepackage{amsmath}
\usepackage{amstext}
\usepackage{amssymb}
\usepackage{amsfonts}
\usepackage[amsmath,thmmarks]{ntheorem}
\usepackage{listings}

\newtheorem{definition}{Definition}
\newtheorem{proposition}{Proposition}
\newtheorem{example}{Example}
\usepackage[inline]{enumitem}
\usepackage{wrapfig}

\usepackage{algorithm2e}
\usepackage{color}
\usepackage{url}

\newif\ifdotikz\dotikztrue

\ifdotikz
\usepackage{pgf}
\usepackage{tikz}
\usetikzlibrary{arrows,positioning,automata,decorations,fit,backgrounds}
\usepgflibrary{shapes.geometric} 
\usetikzlibrary{shapes.geometric} 
\usepgflibrary{decorations.pathmorphing} 
\usetikzlibrary{decorations.pathmorphing} 
\usepgflibrary{decorations.text} 
\usetikzlibrary{decorations.text} 
\usetikzlibrary{matrix}
\pgfrealjobname{ms} 
\else
\ifpdf
\long\def\beginpgfgraphicnamed#1#2\endpgfgraphicnamed{\includegraphics{#1}}
\else
\usepackage{epsfig}
\long\def\beginpgfgraphicnamed#1#2\endpgfgraphicnamed{\epsfig{file=#1.eps}}
\fi

\newcommand{\medblacksquare}{\mathbin{\scalebox{0.6}{\ensuremath{\blacksquare}}}} 

\makeatletter%
\@ifclassloaded{IEEEtran}%
{
	\newtheorem{definition}{Definition}
	
}%
{}%
\makeatother%

\renewtheorem{definition}{Definition}
\theorembodyfont{\normalfont}

\theoremsymbol{\ensuremath{\medblacksquare}}
\renewtheorem{example}[examplecounter]{Example}
\theorembodyfont{\normalfont}
\theoremheaderfont{\normalfont\itshape}
\theoremseparator{.~}
\theoremsymbol{\ensuremath{\Box}}

\newcounter{myenumctr}

\newcommand{\window}{\ensuremath{\boxplus}}

\newcommand{\mi}[1]{\ensuremath{\mathit{#1}}}

\newcommand{\cA}{\ensuremath{\mathcal{A}}}

\definecolor{dark-gray}{gray}{0.25}

\makeatletter
\lst@Key{countblanklines}{true}[t]%
{\lstKV@SetIf{#1}\lst@ifcountblanklines}

\lst@AddToHook{OnEmptyLine}{%
    \lst@ifnumberblanklines\else%
    \lst@ifcountblanklines\else%
    \advance\c@lstnumber-\@ne\relax%
    \fi%
    \fi}
\makeatother

\lstdefinelanguage{asp}{
    breakatwhitespace=true,
    morecomment=[l]{\%},
    breakatwhitespace=true,
    commentstyle=\it\color{dark-gray},
    captionpos=b, 
    numbers=left,
    numbersep=5pt,
    numberstyle=\tiny\color{dark-gray},
    numberblanklines=false,
    countblanklines=false,
    frame=bt, framexbottommargin=5pt, framextopmargin=5pt,
    aboveskip=5pt, belowskip=5pt,
    abovecaptionskip=10pt
}

\begin{document}
\maketitle

\begin{abstract}
Stream reasoning systems are designed for complex decision-making from
possibly infinite, dynamic streams of data. Modern approaches to
stream reasoning are usually performing their computations using
stand-alone solvers, which incrementally update their internal state
and return results as the new portions of data streams are
pushed. However, the performance of such approaches degrades quickly
as the rates of the input data and the complexity of decision problems
are growing. This problem was already recognized in the area of stream
processing, where systems became distributed in order to allocate vast
computing resources provided by clouds. In this paper we propose a
distributed approach to stream reasoning that can efficiently split
computations among different solvers communicating their results over
data streams.  Moreover, in order to increase the throughput of the distributed
system, we suggest an interval-based semantics for the LARS language, which
enables significant reductions of network traffic. Performed evaluations
indicate that the distributed stream reasoning significantly outperforms
existing stand-alone LARS solvers when the complexity of decision problems
and the rate of incoming data are increasing. 
Under consideration for acceptance in Theory and Practice of Logic Programming.
\end{abstract}


\section{Introduction}\label{sec:intro}
Various applications in emerging domains, such as Cyber-Physical Systems (CPS), Industry Digitalization or Internet of Things, require complex monitoring and decision-making over streams of data.  For instance, in CPS quite often sensors provide data about the environment and decision-making components should use it to timely detect environment changes and determine an appropriate reaction of the system.  Reasoning over streaming data \cite{DBLP:journals/expert/ValleCHF09,DBLP:journals/vldb/ArasuBW06,DBLP:conf/dagstuhl/HeintzKD10}
is a recently emerged logic-based paradigm that allows for a declarative representation and solving of various problems. The Logic-based framework for Analytic Reasoning over Streams (LARS) \cite{DBLP:journals/ai/BeckDE18} extends Answer Set Programming (ASP) with window operators and time modalities to declaratively specify complex decision problems on streaming data.

Reasoning in presence of continuously changing data is a challenging problem since a model, returned by a stream reasoner for the previous portion of data, must be updated as soon as possible in order to ensure an acceptable latency/throughput of a streaming system.
Recently a number of \emph{specific stream reasoners} supporting LARS, e.g., Ticker  \cite{DBLP:journals/tplp/BeckEB17} or Laser \cite{DBLP:conf/semweb/BazoobandiBU17}, and \emph{solvers with streaming features}, e.g., \cite{DBLP:conf/esws/BarbieriBCVG10,DBLP:conf/semweb/PhuocDPH11,DBLP:journals/corr/GebserKKS14}, have been proposed.
These reasoners were evaluated in various scenarios which indicated their applicability.

\paragraph{Example (Content caching).}
\begin{wrapfigure}{r}{0.6\textwidth}
	\begin{center}
		\includegraphics[width=0.58\textwidth]{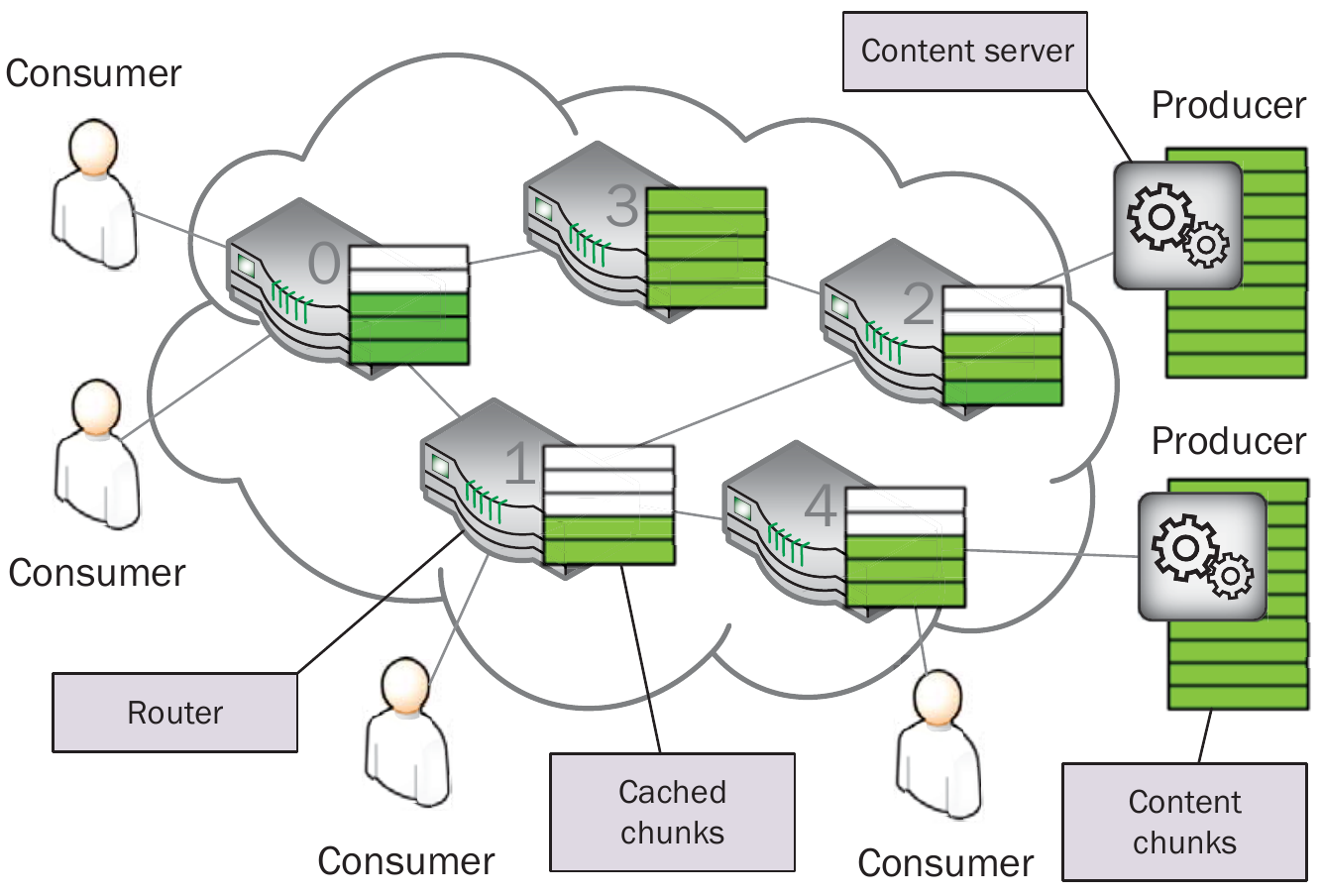}
	\end{center}
	\caption{An example of a CCN architecture \label{fig:example}}
\end{wrapfigure}
	Consider a scenario in which content is distributed over a Content Centric Network (CCN) shown in Fig.~\ref{fig:example}. 
	Content chunks in a CCN can be stored on servers and also in caches of routers.
	If a consumer requests a chunk, it can be delivered from any storage.
	In an ideal situation, a router might satisfy all requests of directly connected consumers from its cache, thus, significantly reducing the network traffic. 
	There are many strategies implemented in CCN routers to decide whether a forwarded chunk should be cached. 
	A router can be configured to use some strategy by network administrators.
	For instance, if some video gets very popular over a short period, an administrator may configure routers to cache highly popular chunks.
	If there is no particular pattern of consumer interests, a random caching strategy might be preferable.
	As it is hard for administrators to monitor user requests in the network, it would be desirable that routers autonomously switch their caching strategies to ensure high quality of service.
\vspace{5pt}

The major advantage of LARS reasoners in such scenarios as described above is that they are able to promptly update previously computed models, whenever the data in a stream is changing.
Instead of computing a new model from scratch, such reasoners apply clever strategies, such as a justification-based truth maintenance system (JTMS) \cite{DBLP:journals/ai/Doyle79} in Ticker or semi-naive evaluation of Datalog programs \cite{DBLP:books/aw/AbiteboulHV95} in Laser, to efficiently update their models.
The solvers with stream features usually cannot hold the update rates of stream reasoners, e.g., evaluation of the ``ticked ASP'' encoding of LARS programs required more time than Ticker \cite{DBLP:journals/tplp/BeckEB17}.
However, they can handle more expressive subsets of LARS and provide more features of the ASP language. 
For instance, Laser accepts only the positive fragment of LARS and
Ticker JTMS cannot evaluate programs with constraints or
odd loops.

In this paper, we suggest a novel approach to stream reasoning employing a distributed architecture that
\begin{enumerate*}[label=\textit{(\roman*)}]
	\item allows for a better performance than specific reasoners supporting the same LARS subset, and
	\item supports all features of the LARS language including integrity constraints.
\end{enumerate*}
Our main contributions are briefly summarized as follows:
	\begin{itemize}[label=--,leftmargin=3mm]
		\item We present a novel interval-based semantics that
	allows for the representation of LARS streams in a compact and
	more suitable form for network communication.
				
		\item Next, we describe a distributed architecture that splits evaluation of LARS programs and maps the obtained parts to a specifically generated network of stream reasoners that use interval streams for communication.
		
		\item Based on the architecture we present a prototypical implementation of a distributed reasoning system that can utilize LARS-solvers, e.g., Ticker, or ASP solvers, e.g., clingo \cite{DBLP:journals/corr/GebserKKS14}, as its reasoners. For ASP solvers supporting external predicates/functions, we suggest an encoding that speed-ups processing of window functions by representing them as queries to the interval stream.
		
		\item Finally, we experimentally compare our approach
	with Ticker on Caching
	Strategy \cite{DBLP:conf/icc/BeckBDEHS17} and evaluate
	scalability of the suggested approach on multiple chained
	n-Queens Completion
	problems \cite{DBLP:journals/jair/GentJN17}. Both  experiments
	demonstrate practical benefits of the distributed system approach.
	\end{itemize}

\section{Interval-based Stream Reasoning}\label{sec:semantics}

\newcommand{\nop}[1]{}
    
We assume an underlying set $\mathcal{A}$ of ground atoms. The
semantics of LARS is defined on streams, which associate 
with each  \emph{time point} $t \in T$ of a \emph{timeline} $T$, i.e.,
an interval $T = [i,j]\subseteq \mathbb{N}$,
a set $\mathcal{A}' \subseteq \mathcal{A}$  of ground atoms. 
In the following, we distinguish \emph{extensional} atoms $\mathcal{A}^E$ for input data and  \emph{intensional} atoms $\mathcal{A}^I$ for derived information.

\begin{definition}[Stream]
A (LARS) \emph{stream} is a pair $S = (T,\nu)$, where 
$T$ is a \emph{timeline} and $\nu: \mathbb{N} \mapsto 2^{\mathcal{A}}$ is an \emph{evaluation function}.
\end{definition}

Informally, $a\in \nu(t)$ means that atom $a$ occurs at time $t$ in
the stream. Following this notion, in a distributed system at every
time $t$ a set $\nu(t)$ of atoms must be communicated to its components, even if it did not change, i.e., $\nu(t) = \nu(t{-}1)$. Such communication
might cause a significant load on the networking infrastructure and
thus decrease the overall performance of the system. It is therefore more
economical to report only changes of the atom occurrence. Furthermore,
occurrences may be reported by different observers and appear in a
single stream. We thus adopt in this work a new view of LARS semantics
based on interval streams. For ample background on LARS, see \cite{DBLP:journals/ai/BeckDE18}.

\begin{definition}[Interval stream]
    \label{def:interval-stream}
An \emph{interval stream}\/ is pair $S_I=(T,\eta)$, where
$T$ is a timeline and $\eta: \mathcal{A}\rightarrow 2^{\mathcal{I}(T)}$
is a mapping that assigns to every atom $a \in \mathcal{A}$ a subset 
of the set of nonempty closed intervals over $T$, denoted by
$\mathcal{I}(T) = \{ I= [i,j] \mid I \subseteq T\}$.
\end{definition}

Informally, each interval $I=[i,j]$ in $\eta(a)$ is a record of an occurrence
observation for the atom $a$ at each time point $t\in I$, where $i$ is the begin 
and $j{+}1$ the end of the observation. LARS streams amount to the
special case where all intervals $I \in \eta(a)$ are point intervals $I=[t,t]$.
Note that interval streams are defined only over closed intervals,
as allowing open intervals might result in unjustified
derivations over intervals that extend over the current time point.
A stream reasoner must thus close all open intervals at the time point at which reasoning occurs.

\paragraph{Example (cont.).}
\begin{figure}[ht]
	\centering
	\includegraphics[width=\linewidth]{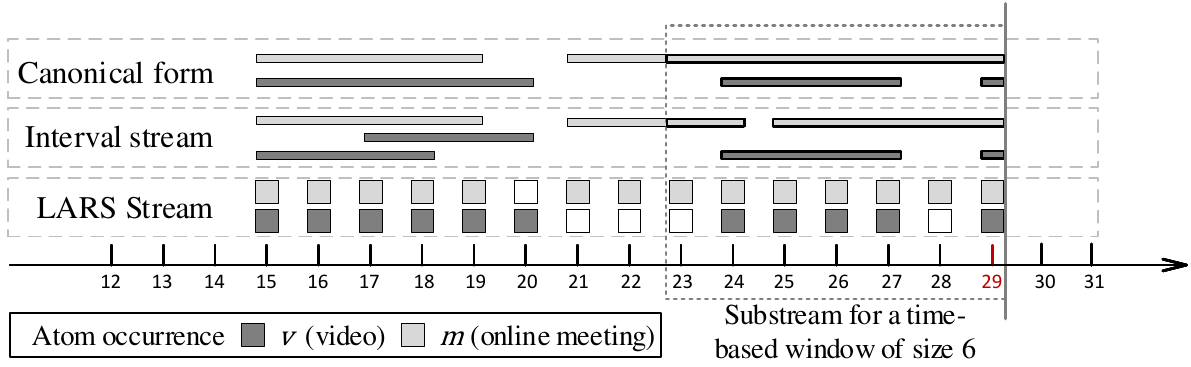}
	\caption{Sample timeline for the Content caching example at the evaluation time point 29}
	\label{fig:streams_example}
\end{figure}
A typical CCN router has a number of networking ports that can receive consumer requests and deliver required data. 
Consider a situation, shown in Fig.~\ref{fig:streams_example}, in which the interval stream represents consumer requests that were registered by three ports and associated with one of the two atoms denoting a music video and an online meeting.
Thus, consumers have been sending requests for the same music video to port 1 between time points 15--19 and 24--27, and port 2 in the interval 17--20.
In addition, port 3 served requests from a customer participating in an online meeting.
Communication of the interval stream generated from these observations requires sending only 
\begin{enumerate*}[label=\textit{(\roman*)}]
	\item the time point at which the first request for the video was registered by a port observer and
	\item the time point at which these requests were no longer observed.
\end{enumerate*}
The LARS stream shown below would send all received observations at every time point thus increasing the load on the network infrastructure.
Moreover, if the information about origins of intervals is not required, stream reasoners can transform interval streams to a canonical form that might result in further reductions of the network traffic, as discussed below. %
\vspace{5pt}

As mentioned above, the occurrence reports at every time point should be semantically merged; 
this leads to the following notion of equivalence.
\begin{definition}[Equivalent interval streams]
Two interval streams $S_I=(T,\eta)$ and
$S'_I=(T',\eta')$ are \emph{equivalent}, if $T=T'$ and 
for every $a\in \mathcal{A}$, $\bigcup \eta(a) = \bigcup \eta'(a)$.
\end{definition}

\noindent Clearly,

\begin{itemize}
\item every interval stream $S_I(T,\eta)$ is equivalent to a unique
LARS stream $S = (T,\nu)$ where $\nu(t) = \{a \in \mathcal{A} \mid
        t \in \bigcup \eta(a) \}$ for all $t \in T$, and 
\item   every LARS stream $S=(T,\nu)$ is equivalent to a unique
interval stream $S'_I=(T,\eta')$ such that for every atom
        $a\in \mathcal{A}$ and intervals $I\neq I' \in \eta'(a)$ it holds that
        $I \cap I'= \emptyset$ and $I\cup I'$ is not an interval.
\end{itemize}
The mapping $\eta'$ can be constructed as $\eta'(a)= \{[t_1, t_2] \subseteq T \mid  [t_1,t_2] = [t_1{-}1, t_2{+}1] \cap \bigcup\eta(a)\}$.
We call $S'_I$ also the \emph{canonical form} of $S_I$, denoted $can(S_I)$.

To deal with the large volume of the streaming data and to simplify
the reasoning tasks, it is useful to consider only substreams of
the original stream. Formally,

\begin{definition}[Substream]
Given interval streams $S_I = (T,\eta)$ and $S'_I = (T',\eta')$,
        $S'_I$ is a \emph{substream} of $S_I$, denoted $S_I'\subseteq
        S_I$, if 
        $T' \subseteq T$ and 
        for every $I' \in \eta'(a)$, where $a \in \mathcal{A}$,
        some $I \in \eta(a)$ exists such that $I'\subseteq I$.
\end{definition}
That is, each occurrence observation in a substream is covered by one
in the original stream. As in LARS, retrieval of a substream can be done by a window function $w$, which can be accessed in rules by a window operator $\window^w$.

\begin{definition}[Window function]
A \emph{window function} $w$ is any (computable) function that given
an interval stream $S_I = (T,\eta)$ and a time point $t$, returns a
substream $S'_I = w(S_I,t)$ of $S_I$.
\end{definition}
Popular in stream processing are sliding \emph{time-based} and \emph{tuple-based} window functions.
Roughly speaking, a time-based window of length $n$ clips the stream to the interval $I_w = [t',t]$, where
$t,t' \in T=[l,u]$, $t$ is the evaluation time point and
$t'= \text{max}(l,t{-}n)$; a tuple-based window of size $n$ 
clips the stream to a shortest interval $I_w = [t',t]$ such that $n$ 
atom occurrences are in it. We omit formal definitions of specific window
operators here, which can be made
following \cite{DBLP:journals/ai/BeckDE18}.

\paragraph{Example (cont.)}
An example of a sliding time-based window of the size 6, denoted $\window^6$, is shown in Fig.~\ref{fig:streams_example}. Given the evaluation time point $t=29$, this function returns a substream with the timeline $T=[23,29]$. The interval $[21,29]$, corresponding to the occurrence of the atom $m$, is clipped at the time point $23$.
\vspace{5pt}

In addition to window operators, time in LARS can also be referred
with \emph{temporal modalities}, 
viz.\ the \emph{at} operator $@_t$, the \emph{everywhere} operator $\Box$, and the \emph{somewhere} operator $\Diamond$. 

The set of \emph{streaming atoms} $\mathcal{A}^+$ is defined by the grammar:

 \smallskip
  
\centerline{$a \mid @_{t}a \mid \window^w @_{t}a \mid \window^w \Diamond a \mid \window^w \Box a$,}
 
 \smallskip
 
\noindent where $a \in \mathcal{A}$ and $t$ is any time point in $T$.

Satisfaction is defined as follows: Given a structure $M=(S_I,W)$ where $S_I=(T,\eta)$ is an interval stream and $W$ a set of window functions $w$ associated with window operators $\window^w$, and the evaluation time point $t \in T$, we have
\begin{itemize}
\item atom $a \in \mathcal{A}$ holds, if
    $t \in \bigcup\eta(a)$, i.e., some $I \in \eta(a)$ exists such that $t\in I$;
	
\item $\Box a$ holds, if $\bigcup\eta(a) = T$,
i.e., for every $t' \in T$ some $I \in \eta(a)$ exists such that $t' \in I$;
	
 \item $\Diamond a$ holds, if  $\bigcup\eta(a) \neq\emptyset$,
i.e., some $I \in \eta(a)$ exists such that $t \in I$;
\item $@_{t'} a$ 
if     $t' \in \bigcup\eta(a)$, i.e., some $I \in \eta(a)$ exists such that $t'\in I$; and 
 \item $\window^w \alpha$ holds, if $\alpha$ holds for $w(S_I,t)$ at $t$.
\end{itemize}

\paragraph{Example (cont.)}
Consider an interval stream $S_I$ over the timeline $[12,31]$ shown in
Fig.~\ref{fig:streams_example}. At the evaluation time point $t=29$,
streaming atoms $\window^6\Box m$, $\window^6\Diamond v$, and $@_{t-3}
v$ hold, whereas $\window^6 @_{28} v$ and $\Box m$ do not hold.

\paragraph{Plain LARS.} A plain LARS program $\Pi$ is a set of rules of the form 
$$r: \quad \alpha \gets \beta_1,\dots,\beta_m,\text{not } \beta_{m+1}, \dots, \text{not } \beta_n$$
where $\alpha$ is a streaming atom 
$a$ or $@_t a$
and the $\beta_i$'s are streaming atoms; $H(r) = \alpha$ is the head of $r$  and
$B(r)=B^+(r)\cup \text{not } B^-(r)$ the body,
where $\text{not } S = \{ \text{not } \beta \mid \beta \in S    \}$ and
$B^+(r)=\{\beta_1,\dots,\beta_m\}$ resp.\
$B^-(r)=\{\beta_{m+1},\dots,\beta_n\}$ is the positive resp.\
negative body.

Given a structure $M = (S_I,W)$ where $S_I=(T,\eta)$
is an interval stream and  $W$ a set of window functions, the evaluation time point $t \in T$, we define that:
\begin{itemize}
	\item $M,t \models \alpha$, where $\alpha \in \mathcal{A}^+$, if $\alpha$ holds in $S_I$ at $t$;
	\item $M,t \models B(r)$ if $M,t \models \beta_i$ and $M,t \not\models \beta_j$, where $\beta_i \in B^+(r)$ and $\beta_j \in B^-(r)$;
	\item $M,t \models r$ if $M,t \models B(r)$ implies $M,t \models \alpha$;
	\item $M,t \models \Pi$ if for every rule $r$ in $\Pi$ it holds that $M,t \models r$.
\end{itemize}
We call $M$ a \emph{model} of $\Pi$ at $t$, if $M,t \models \Pi$. We
note that modelhood is  independent of the syntactic form of an
interval stream, provided that window functions are independent as well. 
We call a window function $w$ \emph{equivalence-preserving}, if for any equivalent
interval streams $S_I$ and $S'_I$ and time $t$, $w(S_I,t)$ is
equivalent to $w(S'_I,t)$. E.g. time-based sliding window functions have this property.

\begin{proposition}
Let $M =(S_I,W)$ be a structure such that every $w\in W$ is
equivalence-preserving. Then for every $S'_I$ equivalent to $S_I$,
streaming atom (resp.\ rule, program) $\Phi$, and time point $t$, it
holds that $M,
t\models \Phi$ iff $M'=(S'_I,W), t \models \Phi$.
\end{proposition}

\paragraph{Answer Streams.}
We introduce the notion of 
\emph{data stream}, which are interval streams
$D_I=(T,\eta_D)$ such that $\eta(a) = \emptyset$ for every atom $a \in \mathcal{A} \setminus
\mathcal{A}^E$, where $\mathcal{A}^E$ is a designated set of
extensional (data) atoms, which represent, e.g., sensor data, input from other systems, etc.
An interval stream $S_{\mathit{Int}} = (T, \eta)$
is then an \emph{interpretation stream} for $D_I$, if $\eta(a) = \eta_D(a)$, for all
$a \in \mathcal{A}^E$.

Answer streams are now defined as follows. We assume that the
interpretation  $W$ of window functions is fixed and identify 
in the sequel structures $(S_I,W)$ with streams $S_I$.

\begin{definition}[Answer Stream]
Given a data stream $D_I=(T,\eta_D)$ and a plain LARS program $\Pi$, an interpretation stream
$S_{\mathit{Int}}=(T,\eta)$ for $D_I$ is an answer set of
$\Pi$, if 
\begin{enumerate*}[label=\textit{(\roman*)}]
	\item $S_{\mathit{Int}}, t \models \Pi$, and
	
	\item every substream $S'_{\mathit{Int}}$ of $S_{\mathit{Int}}$ that is an
	interpretation stream for $D_I$ such that $S'_{\mathit{Int}},t \models \Pi^{S_{\mathit{Int}},t} = \{r\in \Pi \mid S_{\mathit{Int}},t \models
	B(r)\}$ is equivalent to $S_{\mathit{Int}}$.
\end{enumerate*}
\end{definition}

The definition of semantics generalizes the answer stream semantics of 
LARS. Denote for
a LARS stream $S=(T,\nu)$ by $I(S)=(T,\eta)$ the interval stream such
that $\eta(a) = \{ [t,t] \mid t \in T, a \in \nu(t) \}$ for all atoms $a$.

\begin{proposition}
A (LARS) stream $S$ is an answer stream of a plain LARS program $\Pi$
for a (LARS) data stream $D$ 
iff $I(S)$ is an answer stream of $\Pi$ for $I(D)$ under the interval semantics 
, provided that the window functions $W$ preserve equivalence.
\end{proposition}
Notably, for the canonical form this holds in fact for any
interpretation stream $S_I$ for a data stream $D_I$ in place of $I(S)$
resp.\ $I(D)$; we exploit this for distributed computation.

\section{Distributed Architecture}\label{sec:system}
Distributed reasoning for LARS is based on the fact that the evaluation of streaming atoms over different substreams might be performed sequentially by interconnected stream reasoners.
The idea of stream stratification was presented by \citeN{DBLP:conf/ijcai/BeckDE15} and has the same rationale as the classic stratification of logic programs with acyclic negation \cite{DBLP:books/mk/minker88/AptBW88}.
However, instead of searching for strata wrt.\ negation, stream stratification aims at splitting LARS programs along the stream into layers such that computation of window functions in each layer depends only on evaluation results of previous layers.
Obviously, this is possible if the input program has no recursion through window operators. 
Given a set of layers a distributed reasoning system can generate a network of components which can process layer separately and use all possibilities of modern multicore processors and server clusters.

Computation on stream strata is based on the \emph{stream dependency graph} of a LARS
program which allowing detection of recursion over window operators and computation of stream strata.
In this work, we extend the original concept by including negation as a dependency, intuitively, to ensure that choices are made before (or when) they are used.

\begin{definition}[Dependency Graph, \citeNP{DBLP:conf/ijcai/BeckDE15}]
    \label{def:depgraph}
    The (stream) \emph{dependency graph} (wrt.\ time and negation) of a LARS program
    $\Pi$, given as $G_\Pi = (V,E)$, where  
    \begin{enumerate}[label=\emph{(\arabic*)},leftmargin=8mm]
    	\item $V \subseteq \cA^+$ is the set of all streaming atoms occurring in $\Pi$;
    	\item $E$ is a set edges corresponding to relations between atoms, i.e., 
    	for $u,v \in V$, $E$ consists of edges $u\ {rel}\ v$,
    		where $u,v \in E$ and $rel \in \{ \geq, >, = \}$, as follows:

    \begin{enumerate}
        \item if $(u \leftarrow \beta) \in \Pi$ and $v \in \beta$, then $(u \geq v) \in E$;
        \item if $u \in \{\boxplus^w @_t v, @_t v\}$, then $(u = v), (v = u) \in E$;
        \item if $u \in \{\boxplus^w \Box v, \boxplus^w \Diamond v\}$, then $(u > v) \in E$;
            \item if $u \in B^-(r)$ for some $r \in \Pi$, then $(u \geq \text{not}\,u) \in
        E$  (i.e., $v = \text{not}\,u$).
    \end{enumerate}
    \end{enumerate}
\end{definition}
Items (a)--(c) in Definition~\ref{def:depgraph} correspond to the stream
dependency graph in the original paper and item~(d) adds negation. 
Note that this dependency graph differs from the classic one
for logic programs in which \citeN{DBLP:books/mk/minker88/AptBW88} define edges for head to body references that are regarded negative for
body atoms under negation. Our definition creates a dependency between the atom and its own negation.

The $>$-labelled edges in the dependency graph define temporal boundaries in the
reasoning process. The following definition captures the partitioning of a LARS
program into subprograms.


\begin{definition}[Component Graph]
	\label{def:comp_graph}
The \emph{component graph} $C_\Pi = (V,E)$ is defined by construction from $G_\Pi$

    \begin{enumerate}[leftmargin=8mm]
     \item Determine connected components $Comp$ in $G'_\Pi$ obtained from $G_\Pi$ by removing all $>$-labeled edges and considering all edged as undirected.
     
     \item For each $c \in Comp$, label $c$ with the set of rules that
           derive the atoms in this component and add $c$ as node to $V$.
           
     \item For each $>$-labeled edge $e = (a',b')$ in $G_\Pi$, add
            $(a,b)$ to $E$ if $a' \in a$ and $b' \in b$.

     \item Add a special master node $m$ that represents the
          outside world  to $V$ and edges $(m,v)$, $(w,m)$ to $E$ for
          all source nodes $v$ and sink nodes $v$ of the graph from 1--3.
    \end{enumerate}
\end{definition}

\citeN{DBLP:conf/ijcai/BeckDE15} call a program $\Pi$ \emph{stream-stratified}, if a mapping $\lambda:\mathcal{A}^+\rightarrow \{0,\ldots,n\}$, $n\geq 0$, exists s.t.\ $\lambda(u)\ rel\ \lambda(v)$ whenever $G_\Pi$ has an edge $u\ rel\ u$. Clearly, the following result holds.
\begin{proposition}
	A LARS program $\Pi$ is stream-stratified iff
	every cycle of $C_\Pi$ contains the master node.
\end{proposition}

We view the component graph as a network, where a stream reasoner is running in every node. 
Each reasoner takes inputs from the network as defined by its
incoming edges in $C_\Pi$ and distributes its answer stream into the network,
along its outgoing edges in the graph. 
To reflect this, we label each node with different sets of atoms as follows.

\begin{definition}[Component Graph Labeling]
    \label{def:labelsets}
	 Let $C_\Pi = (V,E)$ be a component graph. Then for each $u\in V$ and $(u,v) \in E$, let $pub(u)$, $listening(u,v)$ and $augwant(u)$ be the smallest sets of atoms such that%
    \begin{itemize}[leftmargin=8mm]
        \item $pub(u) = \bigcup_{(u,w) \in E} augwant(w)$
        \item $listening(u,v) = pub(u) \cap augwant(v)$
        \item $augwant(u) = want(u) \cup \left( pub(u) - prod(u)\right)$
    \end{itemize}
where 
$want(u)=\{a \in \cA \mid \forall r \in u:\; a \in B^+(r) \cup B^-(r) \}$, $prod(u)=\{a \in \cA \mid \forall r \in u:\; a \in H(r) \}$, and for $u=m$ (i.e.\ the master node)
$augwant(m) = want(m) = \mathcal{A}^I$, and $pub(m) =prod(m) = \mathcal{A}^E$.
\end{definition}

\begin{figure}[t]
	\centering
	
	\include{figures/mmedia-both}
	
	\caption{Content Caching example in LARS (left) and labeled component graph (right)} \label{fig:caching-both} 
	\label{fig:caching-ticker}
	\label{fig:caching-compgraph}
\end{figure}

The intuition behind the definitions is as follows. 
The \emph{pub}\/ set describes the set of atoms that a node publishes to any of its successors. 
The \emph{augwant}\/ (augmented want) set are the atoms that a node requires from its predecessors to perform its own computations, as well as those it must pass on to its successors. 
Finally, we assert that the master node publishes all extensional atoms and wants all intensional atoms.
Since $augwant(m) = want(m)$, the \emph{passthrough} set, which is implicit in
the definition of the augmented want set, is empty. In other words, we assert
that the master node does not distribute inferences back into the network. 

\paragraph{Example (cont.)} Consider the LARS encoding of the Content Caching problem shown in Fig.~\ref{fig:caching-both}. According to the Definitions~\ref{def:depgraph} and \ref{def:comp_graph} the program can be split into two components including rules $r_1,r_2, \text{ and }r_3$ as well as rules $r_4$ to $r_{10}$, resp. The first component in this case works as a monitor overlooking the distribution of content in the network, i.e., $\alpha$-value \cite{DBLP:conf/icc/BeckBDEHS17}, whereas the second component makes decisions about the strategy. The edges of the component graph are labeled according to Definition~\ref{def:labelsets}.
\vspace{5pt}

\begin{figure}[tb]
	\centering
	\includegraphics[scale=0.85]{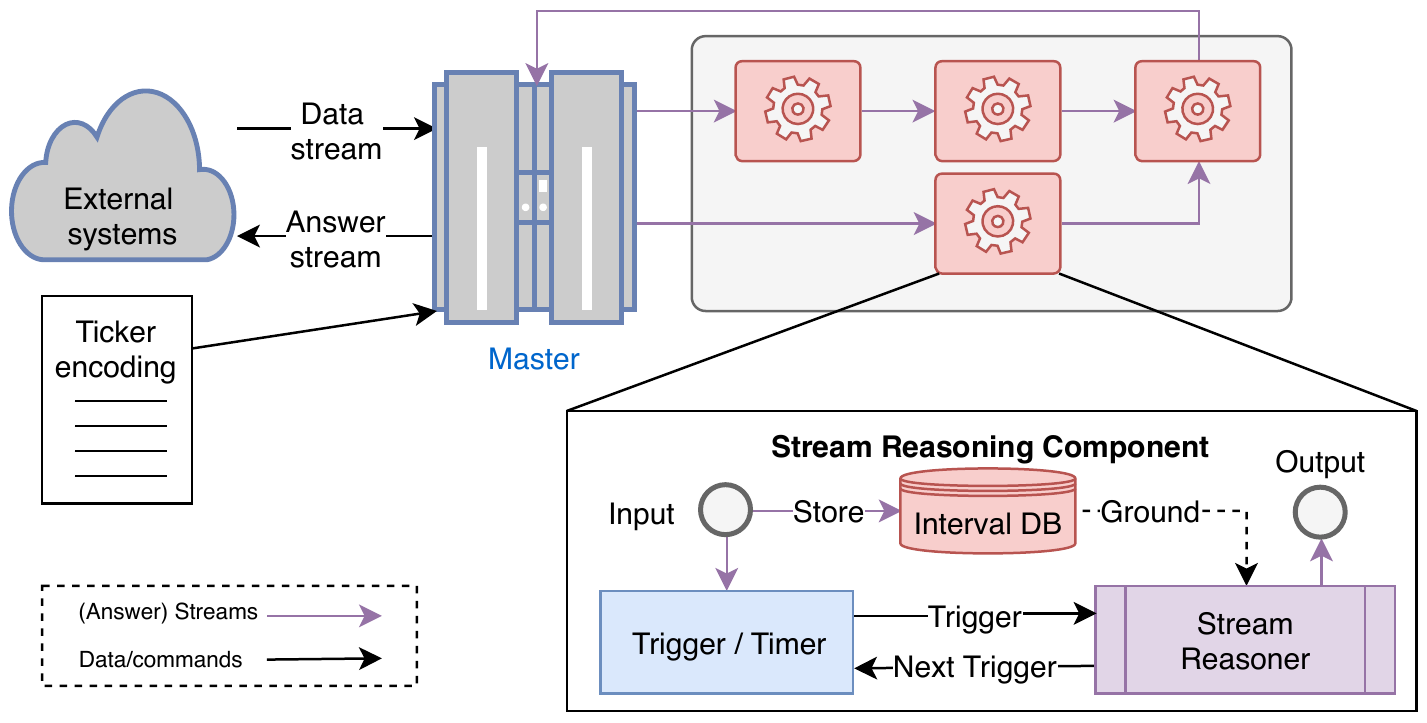}
	\caption{Architecture Overview}
	\label{fig:architecture}
\end{figure}

Figure~\ref{fig:architecture} shows the overall architecture of our system. The master node computes the component graph and spawns nodes in the network
according to the computed topology, shown as the area shaded in gray.
The master is connected to the outside world,
receiving a data stream and supplying answers back. 
Each spawned node is a stream reasoner in charge of the portion of the input program designated by the component graph.

\section{Prototype Implementation} 
The architecture presented so far is reasoner agnostic. We now present a reasoner component used in our prototype, which utilizes an ASP solver
with a reduction of LARS to ASP. 
The cutout in Fig.~\ref{fig:architecture} shows the architecture of a reasoner component. Incoming data stream is converted into the canonical form and stored in the local interval database.
Similarly to previous stream reasoning systems, incoming data triggers the reasoner that computes an answer stream and provides it to the network according to $C_\Pi$. 

Since ASP systems cannot process LARS programs directly, we use a LARS
to ASP encoding that replaces all streaming atoms with
an \emph{external function} $f$ that returns $1$, if the provided
condition is satisfied over the canonical interval stream stored in the database $\mi{DB}$,  and $0$ otherwise.
\[
\begin{array}{rll}
\boxplus^k \square a &\rightarrow& 1 = f(\forall t : t \in (N - k, N) \rightarrow \text{$a$ is true at $t$ in $\mi{DB}$})\\
\boxplus^k \Diamond a &\rightarrow& 1 = f(\exists t : \text{$a$ is true at $t$ in $\mi{DB}$} \wedge t \in (N - k, N))\\
\boxplus^k @_T a &\rightarrow&  \{T = t : \text{$a$ is true at $t$ in $\mi{DB}$} \wedge t \in (N - k, N)\}\\
@_t a &\rightarrow& 1 = f(\text{$a$ is true at $t$ in $\mi{DB}$})\\
\end{array}
\]
Applying external functions \cite{DBLP:journals/corr/GebserKKS14} allows for significant performance improvements during grounding. 
To store the canonical interval stream and evaluate the function $f$ efficiently, we implement a custom in-memory database for intervals. 
The incoming stream is processed by the \textsc{Occurrences}$(L)$ function that takes a list $L=(L^+, L^-)$ of atom occurrences and disappearances, resp., and updates the database. 
The function
assumes that observations remain as they are
unless changed and that all external sources report about
changes reliably.
The \textsc{Cleanup}$(t)$ function is called periodically to purge parts of the interval stream that can safely be forgotten; the cutoff point $t$ can be determined by static program analysis.
Note that the implementation of the \textsc{Occurrences} function closes and opens intervals always with the current evaluation time point $t$. 
To track atoms associated with future time points, e.g.,
$@_{t+k}a \leftarrow \beta$ for $k > 0$, the master node keeps 
all such future inferences in a special log
and puts them back into the input stream at the appropriate time. 

Previous LARS systems have typically utilized a steady tick stream, such that
reasoning is performed repeatedly, even in the absence of ``new''
information. Without such a periodic reasoning trigger, computed answer
streams may be incorrect.

\paragraph{Example (cont.)} Suppose the rule
$\mi{high} \leftarrow \boxplus^6 \square v$ is evaluated against the
interval stream shown in Fig.~\ref{fig:streams_example}. At the
evaluation time point $t=29$, this rule cannot be used to derive
$\mi{high}$. If no new data appears in the reasoner's input, then at the evaluation time point $t=35$ the atom $\mi{high}$ can actually be derived. 
However, as no new data occurred, the reasoner will not be triggered and therefore the answer stream will not be changed. 
\vspace{5pt}

The Trigger/Timer component in Fig.~\ref{fig:architecture} keeps track of the next windows that must be reevaluated in the absence of new data and triggers the reasoner when required. 
In particular, for streaming atoms of the form 
$a$, $@_{t}a$, $\window^w @_{t}a$, and $\window^w\Diamond a$, a reasoner can determine whether they hold at the time when new changes in the stream were reported.
The atoms of the form $\window^w \Box a$ are associated by the trigger with a timer that will start the reasoner at an appropriate time, e.g., duration of the shortest time-based window among all atoms or minimal number of ticks at which a tuple-based window might return the required number of tuples.

\section{Evaluation}\label{sec:eval}
We performed two evaluations of our method. The first compared our implementation
on a small example against the current state-of-the-art LARS reasoner, Ticker.
The second experiment seeks to examine the speedup attainable in a fully
utilized pipeline of reasoners. In the following, we refer to proper
Ticker, using the incremental reasoning procedure, as Ticker-JTMS and
to the clingo-based reasoner implemented in Ticker as Ticker-ASP.
Analogously, refer
to our system equipped with Ticker in all reasoner nodes (resp.\ with our clingo
based reasoner) as Distributed-JTMS (resp.\ Distributed-ASP). 

All experiments were performed on Fedora 30 using an AMD Ryzen 2700X, and
16 GiB of memory. For more information on the experiments and replication
thereof, please see the project
website,\footnote{\url{http://distributed-stream-reasoner.ainf.at}}
which 
will be amended with additional experiments.

\subsection{Reasoner Performance}
For our first experiment, we reused the Content Caching example in
Fig.~\ref{fig:caching-ticker}, which had been used in \cite{DBLP:journals/tplp/BeckEB17}. 
Similar to the evaluation of
Ticker, we vary two parameters: the size of the windows used (named $k$ in
Fig.~\ref{fig:caching-ticker}), and the length of the data stream.
However, our implementation does not support tuple-based windows at the moment, we used time-based windows instead. 
Our data stream is regularly spaced out in time, and our windows are sized such that the same number of atom occurrences fall into each window as in the evaluation by \citeN{DBLP:journals/tplp/BeckEB17}. 
We also repeated the evaluation on Ticker, also using time-based windows, to control for differences in hardware used. 
The incoming data stream is spaced at 0.1 seconds intervals and every batch of data comprised \texttt{alpha} values such that all use cases of the problem are activated periodically. 
This means that the reasoners ideally have some idle time.
We therefore measure the CPU time of each run, which we obtain from the
Linux kernel through the \texttt{/proc} file system.

\begin{figure}[t]
    \centering
    \includegraphics[scale=0.27]{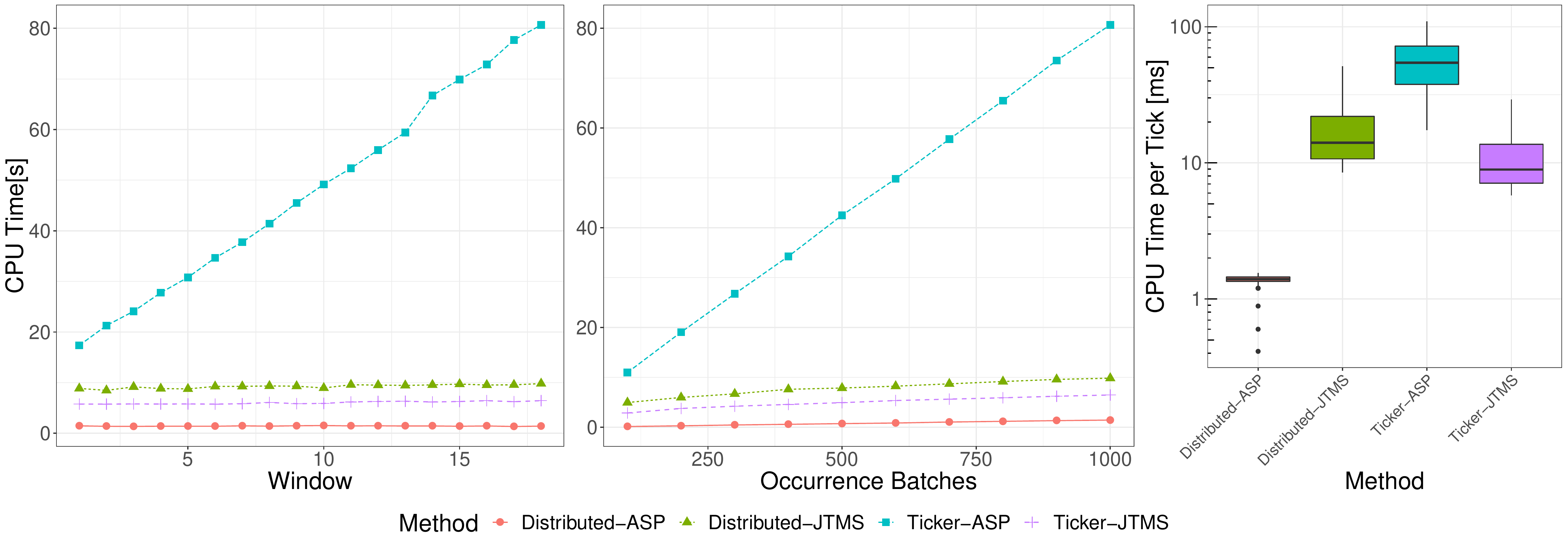}
    \caption{Total CPU Time Evaluation on the Video Caching Problem.}
    \label{fig:ticker-comparison}
\end{figure}

Figure~\ref{fig:ticker-comparison} shows the results of this experiment. 
On the left, the stream length is held constant at 1000 ticks and the
window size is increased along the x-axis. 
It can be seen that neither Ticker-JTMS nor our method is affected by window size. Ticker-ASP scales linearly with the window size, confirming the experimental results of \citeN{DBLP:journals/tplp/BeckEB17}.

In the middle, we hold the window size constant at $k = 18$, and scale the
number of data occurrences along the x-axis. The chart on the right-hand side shows a comparison of methods on a per tick basis. Note that in this chart, the y-axis is scaled logarithmically.

Overall the Distributed-ASP approach outperforms the other methods.
The difference between Ticker-JTMS and
Distributed-JTMS is noteworthy. Since Ticker always ``ticks'', the distributed
system requires roughly twice as much CPU time as the non-distributed approach.
Distributed-ASP on the other hand practices a kind of ``lazy evaluation''. The
second reasoner node will not perform any work unless it receives new inputs,
which only happens whenever the first node produces a different answer stream,
containing literals relevant to the second node. As a result, the second
reasoner node performs much less work in Distributed-ASP than
in Distributed-JTMS.

In our evaluation, a clingo-based system outperforms
Ticker-JTMS, which runs counter to the results obtained by \citeN{DBLP:journals/tplp/BeckEB17}, and validated here as Ticker-ASP. Whereas Ticker-ASP uses a ticked ASP
encoding that can be reused with all ASP solvers, our reasoner implementation is
more specific, utilizing clingo's ability to consult an external oracle during
grounding, but also shows great performance gains.

\subsection{Pipeline Efficiency}

In our second experiment, we benchmarked our system against itself in
different settings. The goal was to examine the extent to which
performance can be increased due to pipeline parallelism. For this purpose we
constructed a program which places non-trivial load on each reasoner and 
results in a long enough chain of reasoners after decomposition.
This is necessary to outweigh the inherent overhead described above,
and is representative of real-world problems. In the first setup, we evaluated
the system in \emph{single node mode}, i.e. the decomposition 
collapses into a single reasoner node. In effect, this corresponds to a
non-distributed system. The second setup leaves the decomposition as is.

The experiment consisted of increasing the speed of the data stream until the
reasoner ``breaks'', determined by increasing latency between the last input
atom and the last output atom. This allows us to determine the point at which
the system saturates, at which it is no longer able to cope with the data
volume.

As the underlying problem, we picked the well-known n-Queens Completion problem \cite{DBLP:journals/jair/GentJN17}, as its difficulty can be easily adjusted to
different needs by varying the parameter $n$. In order to link
multiple such problems, we make subsequent steps temporally
dependent on the configuration calculated in the previous step.
Specifically, the n-Queens problem at step $k$ reuses the placement
of Queen $k-1$ from the prior step via an streaming atom: this scheme
chains multiple n-Queen problems in a row and hence
fulfills the requirements laid out above. The encoding is available in \ref{app:encodings}.


\begin{figure}[t]
    \centering
    \includegraphics[scale=0.27]{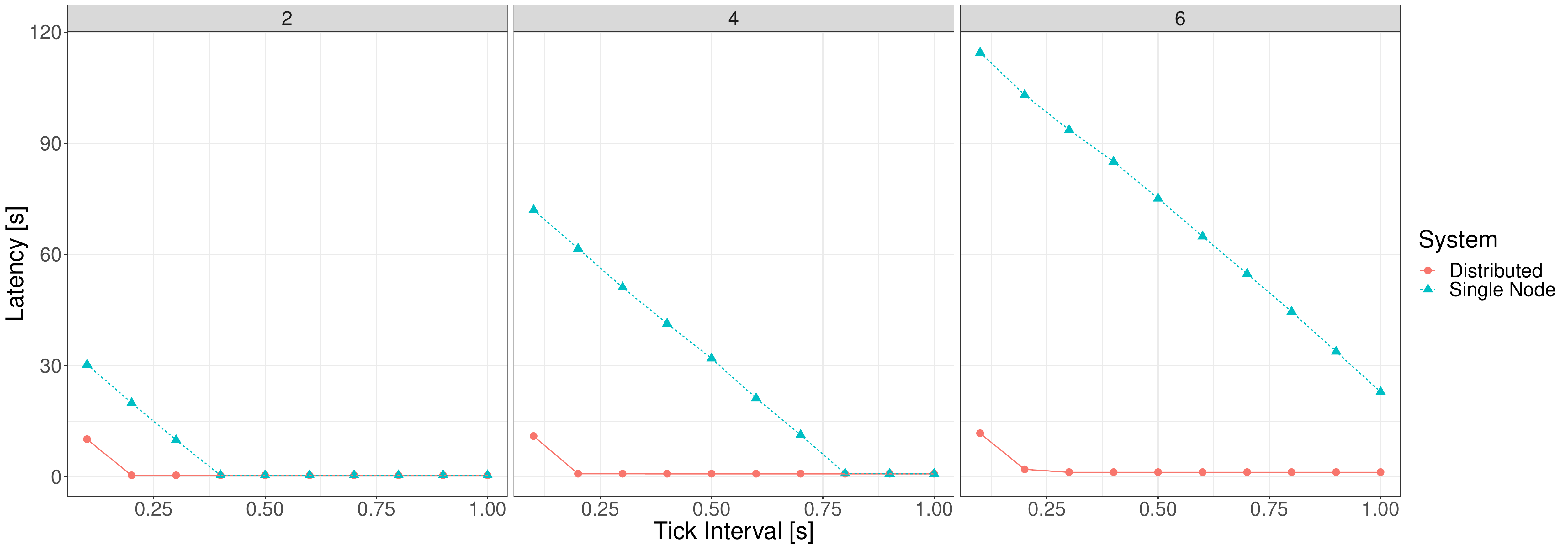}
    \caption{Latency after 100 atom occurrences in the input stream on the N-Queens Problem. 
    From left to right, the graphs depict 2,4,and 6 Pipeline Stages.}
    \label{fig:nqueens-eval}
\end{figure}

Figure~\ref{fig:nqueens-eval} shows the results of the experiment. 
Note that any clear increase in latency in these charts indicates that the system has been saturated and that new data is piling up. 
The specific latency values are due to the atom occurrence rates supplied during the experiment. Given higher rates, the latency would increase even further.

Ticker-ASP was tested but omitted in Figure~\ref{fig:nqueens-eval}, as
no run finished 
within 120 seconds of latency, 
which was the timeout. Ticker-JTMS cannot solve the problem due to the
presence of odd loops, which is an inherent limitation of the
approach \cite{DBLP:journals/tplp/BeckEB17}.

The results show that the distributed version of the system can cope
with higher data volume. In the case of four stages, we observe an
increase in latency 
in the distributed system 
between 0.2 
and 0.1 seconds tick interval, while the single node system shows the
same behaviour between 0.7 and 0.8 seconds tick interval. Splitting
the difference, this is representative of a 
500\% increase in throughput in the distributed over the single node
version. This gets more pronounced as we increase the number of
stages.  The distributed version shows  at six stages similar performance
as at two stages, while the single node version exhibits
increased latency even at a 1 second tick interval for six stages.

\section{Related Work}\label{sec:rw}
Representing time by intervals is used in various logics, e.g.\ 
in Metric Interval Temporal Logic \cite{AlurFH96},
Metric Time Datalog \cite{Brandt17},
or Signal Temporal Logic \cite{DBLP:conf/formats/MalerN04} with applications in CPS.
DyKnow \cite{HeintzD04} was one of the first systems using temporal
logic for continuous and incremental reasoning on streaming data. It
comprises 
components 
that publish and subscribe to stream generators in a processing
network, whose structure is declaratively described.

Furthermore, the interval-based approach is widely used in Complex
Event Processing (CEP) systems, which view data signals as
notifications of events, which are often represented by time
intervals.  ETALIS \cite{DBLP:journals/semweb/AnicicRFS12} is a
notable CEP system that combined CEP techniques, such as pattern
matching and derivation of complex events, with logic programming. Its
rule language has a model-based semantics and allows
to encode complex events using patterns that 
are expressions in  Allen's \citeyear{Allen83} interval algebra.
A rule $a \gets \mathit{pattern}$ 
associates $a$ with any interval that matches 
$\mathit{pattern}$.
As shown in \cite{DBLP:conf/ijcai/BeckDE15,beck-18}, 
ETALIS can be expressed in LARS using a specific window function, 
if the input stream has no overlapping event intervals
for each atom.

Implementations of LARS also considered temporal intervals
for speedups.
In Ticker \cite{DBLP:journals/tplp/BeckEB17}, every atom is associated
with intervals indicating time points in which it has
one of the labels $\mathit{in},$ $\mathit{out},$ $\mathit{unknown}$. 
The latter are used by a
JTMS implemented in Ticker to update the answer stream.
Similarly, Laser \cite{DBLP:conf/semweb/BazoobandiBU17} annotates each
LARS formula $\alpha$ with a substitution and an interval in
which $\alpha$ necessarily holds if no new data arrives. 
%
In all these cases, the intervals
served to avoid unnecessary computations when the arriving data does not affect
(parts of) the current answer stream;
the reasoner can then simply resend previously derived atoms.
%
In this paper, we define LARS semantics using intervals to
foster efficient communication between reasoners.
The ensuing change affects all 
distributed reasoners, as it assumes that triggering model updating
procedures is independent of new data in the input stream,
in contrast to the systems from above.
Nonetheless, both stream reasoners can be used in our architecture if their interval interfaces can be directly accessed by the triggering component.

Distributed architectures are common in stream processing frameworks, e.g., Spark or Flink, but until recently not much considered in stream reasoning. 
The first step towards distributed stream reasoning is presented
in \cite{ren-18,DBLP:conf/semweb/RenCNX18}. This approach suggests two translations of positive LARS programs into BigSR code---an RDF reasoning framework. 
The code obtained is executed on top of Flink or Spark stream processing engines supporting distributed execution.
The approach suggested in this paper implements more expressive (non-stratified) plain LARS programs and uses a number of interconnected stream reasoners for their distributed execution. 

Finally, there are approaches that parallelize computation of answer sets in order to achieve performance improvements allowing for the application of ASP solvers for stream reasoning. 
Thus, \citeN{DBLP:journals/semweb/PhamAM19} suggest an extension to StreamRule framework that uses  a decomposition method to partition an given stratifiable ASP program and the incoming stream of atoms into parts that can be evaluated independently. 
Results obtained by authors show benefits of such parallelization and can definitely be incorporated in our framework to partition programs associated with nodes of the component graph.

\section{Conclusion}
We have presented a distributed approach to stream reasoning that uses
a novel interval-based representation of LARS semantics, aiming to
improve storage and communication performance between reasoning
components.  A given stream-stratifiable plain LARS program is
automatically decomposed into layers whose evaluation
depends only on the results of lower layers.  The layers are mapped to
a generated network of stream reasoners that can process the program
in a distributed way. We have described an implementation of this
architecture in a system that can accommodate various stream reasoners
for the components. The evaluation results indicate that the system
outperforms existing stream reasoners for the same LARS fragment on a
realistic monitoring problem.  Finally, the scalability of the novel
system was shown by a benchmark on a chained n-Queens completion
problem, which could not be solved easily by previous LARS reasoners.

\paragraph{Outlook.}
The work that we have presented can be continued in several directions. On the
one hand, the repertoire of window operators can be enriched and made available
in the implementation; in particular, aggregates that can be viewed as
particular window functions may be provided, but furthermore also added as
first-class citizens to the LARS framework, for both traditional LARS streams
and interval streams.

As regards the distributed computation approach, in particular the following issues are on our agenda:
\begin{enumerate*}[label=\textit{(\roman*)}]
	\item to investigate issues related to distributed computation, such
	as time synchronization or liveness, in order to ensure the
	correctness of answer streams that are output by the system; and
	\item to improve incremental reasoning techniques with on-the-fly
	grounding; and
	\item to provide distributed answer set computation that supports
	full plain LARS, i.e., programs that are not stream-stratified.
\end{enumerate*}

\subsection*{Acknowledgments}
This work was conducted in the scope of the research project \textit{DynaCon (FFG-PNr.: 861263)}, which is funded by the Austrian Federal Ministry of Transport, Innovation and Technology (BMVIT) under the program ``ICT of the Future" between 2017 and 2020 (see \url{https://iktderzukunft.at/en/} for more information).


\newpage
\appendix

\section{Ticker encodings used in the evaluation}
\label{app:encodings}
This appendix presents two Ticker encoding used in the evaluation. The first
encoding is taken without modifications from the evaluation presented in
\cite{DBLP:journals/tplp/BeckEB17}. 

\begin{verbatim}
high :- value(V), alpha(V) at T [3 sec], 18 <= V.
mid :- value(V), alpha(V) at T [3 sec], 12 <= V, V < 18.
low :- value(V), alpha(V) at T [3 sec], V <= 12.
lfu :- high always [3 sec].
lru :- mid always [3 sec].
fifo :- low always [3 sec], rtm50 [3 sec].
done :- lfu.
done :- lru.
done :- fifo.
random :- not done.
value(5). value(15). value(25).
\end{verbatim}

The second encoding for the chained n-Queens completion problem is used in
conjunction with a generator that given the number $k$ of chained problems
outputs an encoding comprising $k$ liked copies of the presented one. All
predicates in every encoding $i$ of the chain are renamed according to the
value of $i$ for every sub-problem. Every of the generated problems corresponds
to the 18-Queens problem.
 \begin{align*}
 	 d_i(1..n).& \\
     q_i(i,X) &\gets \boxplus^w \Diamond \mi{send}_{i-1}(X).\\
     q_i(R,C) & \gets d_i(R), d_i(C), \mathit{not}~ \mathit{not}~ q_i(R,C). \\
     & \gets q_i(R,C), q_i(R1,C), R<R1. \\
     & \gets q_i(R,C), q_i(R,C1), C<C1. \\
     & \gets q_i(R,C), q_i(R1,C1), R<R1, R1-R = |C1-C|.\\
     p_i(R) &\gets q_i(R,C).\\
     &\gets d_i(Q), \mathit{not}~ p_i(Q).\\
     \mi{send}_i(X) & \gets q_i(i,X).
 \end{align*}

\noindent Ticker version of the LARS chained n-Queens encoding for $i=1$ is shown below.

\begin{verbatim}
q1(1,Y) :- send0(Y) in [50 msec], d1(Y).
d1(1..18).
 q1(X,Y) :- d1(Y), d1(X), not -q1(X,Y).
-q1(X,Y) :- d1(Y), d1(X), not  q1(X,Y).
:- q1(X1,Y), q1(X2,Y), X1 < X2.
:- q1(X,Y1), q1(X,Y2), Y1 < Y2.
:- q1(X1,Y1), q1(X2,Y2), X1 < X2, Y1 < Y2, X2 - X1 = Y2 - Y1.
:- q1(X1,Y1), q1(X2,Y2), X1 < X2, Y2 < Y1, X2 - X1 = Y1 - Y2.
placed1(X) :- q1(X,Y), d1(Y).
:- not placed1(X), d1(X)
send1(Y) :- q1(1,Y).
#show send1/1.
\end{verbatim}

\end{document}